\pdfoutput=1
\documentclass[11pt]{article}

\usepackage[final]{coling}

\usepackage{times}
\usepackage{latexsym}
\usepackage{multirow}
\usepackage{amsmath}
\usepackage{arydshln}
\usepackage{float}
\usepackage{amssymb}
\usepackage[T1]{fontenc}
\usepackage[utf8]{inputenc}

\usepackage{microtype}

\usepackage{inconsolata}

\usepackage{graphicx}

\title{FuocChuVIP123 at CoMeDi Shared Task: Disagreement Ranking with XLM-Roberta Sentence Embeddings and Deep Neural Regression}

\author{Chu Duong Huy Phuoc \\
  University of Information Technology \\
  Vietnam National University - Ho Chi Minh City, Vietnam \\
  \texttt{23521229@gm.uit.edu.vn} 
}
\begin{document}
\maketitle
\begin{abstract}
This paper presents results of our system for CoMeDi Shared Task, focusing on Subtask 2: Disagreement Ranking. Our system leverages sentence embeddings generated by the paraphrase-xlm-r-multilingual-v1 model, combined with a deep neural regression model incorporating batch normalization and dropout for improved generalization. By predicting the mean of pairwise judgment differences between annotators, our method explicitly targets disagreement ranking, diverging from traditional "gold label" aggregation approaches. We optimized our system with a customized architecture and training procedure, achieving competitive performance in Spearman correlation against mean disagreement labels. Our results highlight the importance of robust embeddings, effective model architecture, and careful handling of judgment differences for ranking disagreement in multilingual contexts. These findings provide insights into the use of contextualized representations for ordinal judgment tasks and open avenues for further refinement of disagreement prediction models.
\end{abstract}

\section{Introduction}
The CoMeDi Shared Task Subtask 2: Mean Disagreement Ranking with Ordinal Word-in-Context Judgments (DisWiC) \cite{schlechtweg2025comedi} focuses on predicting annotator disagreement in semantic similarity judgments. Participants were tasked to rank word-use pairs based on the mean of pairwise absolute differences in annotations, highlighting disagreement rather than consensus. This task builds on recent research emphasizing the importance of capturing variability in linguistic judgments for complex, ambiguous datasets. Evaluations were using Spearman’s correlation.

In this paper, we present an embedding-based approach that uses SentenceTransformer (paraphrase-xlm-r-multilingual-v1) with base model is XLM-RoBERTa \cite{conneau-etal-2020-unsupervised} to generate contextual embeddings for word-use pairs. These embeddings were combined in a deep regression model with Batch Normalization, Dropout, and an optimized learning rate scheduler to enhance performance. The model was fine-tuned to predict disagreement scores efficiently, demonstrating the potential of leveraging advanced multilingual embeddings and robust neural architectures for capturing semantic complexities in multilingual datasets.

\section{Related Work}
Annotation disagreements in NLP, particularly in tasks involving meaning in context, pose challenges to data quality and model reliability. Early studies, such as \cite{artstein-poesio-2008-survey} and \cite{hovy-etal-2013-learning}, explored inter-annotator agreement and aggregation methods to address inconsistencies. Recent works have shifted toward leveraging disagreements as valuable signals. For instance, \cite{DBLP:journals/corr/abs-2109-04270} introduced perspectivism to embrace diverse annotator viewpoints, while \cite{davani-etal-2022-dealing} and \cite{davani-etal-2022-dealing} utilized disagreements to train models better suited for subjective tasks.
In Word-in-Context (WiC) tasks, \cite{Schlechtwegetal18} proposed the DURel framework to capture semantic relatedness using ordinal scales, with subsequent studies, such as \cite{Uma2021LearningFD}, focusing on preserving disagreement information through alternative label aggregation methods. This Subtask 2 builds on this foundation by explicitly modeling disagreement using mean pairwise judgment differences, evaluated via Spearman’s correlation \cite{Zar2005SpearmanRC}, offering a novel perspective on handling annotation variability.
\section{Task Description}
The CoMeDi shared task, part of the COLING 2025 workshop \cite{schlechtweg2025comedi}, consists of two subtasks focusing on predicting disagreements in word sense annotation in context (WiC). The first subtask (OGWiC) involves predicting the median of annotator judgments on an ordinal scale (1-4) for word usage pairs, treating this as an ordinal classification task. The second subtask (DisWiC) aims to rank instances based on the mean disagreement between annotators, measured by pairwise absolute differences in judgments. Both subtasks rely on datasets such as the DWUG EN dataset \cite{Schlechtweg2024dwugs} and will be evaluated using Krippendorff's \(\alpha\) \cite{krippendorff2018content} for OGWiC and Spearman's \(\rho\) for DisWiC.

\subsection{Dataset}
We conducted our experiments using the dataset provided by the organizers for training and evaluation. The dataset includes samples from seven languages: Chinese \cite{Chen2023chiwug}, English \cite{Schlechtweg2024dwugs}, German \cite{Schlechtweg2024dwugs}, Norwegian \cite{kutuzov2022nordiachange}, Russian \cite{rodina2020rusemshift}; \cite{Kurtyigit2021discovery}, Spanish \cite{Zamora2022lscd}, and Swedish \cite{Schlechtweg2024dwugs}. Tables \ref{tab:dataset} and \ref{tab:dev_dataset} summarize its key characteristics.

The training dataset contains more samples than the development set, ranging from 1,222 for Norwegian to 24,891 for Russian. On average, context length varies widely, with Spanish having the longest at 84.72 tokens and Chinese the shortest at 1.00 token. German has the largest maximum context length of 1,643 tokens, while Chinese remains the smallest at 1 token. This diversity in sample sizes and context lengths across languages poses challenges for model generalization but provides a strong foundation for evaluating multilingual methods.
\begin{table}[h!]
\centering
\resizebox{\columnwidth}{!}{%
\begin{tabular}{lccc}
    \hline
    \textbf{Languages} & \textbf{\# Samples} & \textbf{Avg. Len.} & \textbf{Max Len.} \\
    \hline
    Chinese & 20.46 & 1.00 & 1.00 \\
    English & 10.83 & 31.91 & 176.00 \\
    German & 13.69 & 39.39 & 1643.00 \\
    Norwegian & 6.04 & 47.49 & 346.00 \\
    Russian & 12.69 & 24.88 & 356.00 \\
    Spanish & 9.33 & 84.72 & 480.00 \\
    Swedish & 9.11 & 34.89 & 376.00 \\
    \hline
\end{tabular}%
}
\caption{Training dataset statistics.}
\label{tab:dataset}
\end{table}

\begin{table}[h!]
\centering
\resizebox{\columnwidth}{!}{%
\begin{tabular}{lccc}
    \hline
    \textbf{Languages} & \textbf{\# Samples} & \textbf{Avg. Len.} & \textbf{Max Len.} \\
    \hline
    Chinese & 3.09 & 1.00 & 1.00 \\
    English & 1.90 & 32.01 & 169.00 \\
    German & 2.59 & 33.52 & 376.00 \\
    Norwegian & 871 & 52.89 & 452.00 \\
    Russian & 1,932 & 23.98 & 352.00 \\
    Spanish & 1,269 & 82.19 & 493.00 \\
    Swedish & 1.41 & 33.66 & 305.00 \\
    \hline
\end{tabular}%
}
\caption{Development dataset statistics.}
\label{tab:dev_dataset}
\end{table}

\section{System Overview}
Our system tackles the shared task by combining neural sentence embeddings and a deep regression model to predict mean disagreement rankings for the DWUGs dataset \cite{Schlechtweg2024dwugs}. The primary steps include: 
(i) generating semantic representations using multilingual pre-trained models, 
(ii) concatenating embeddings for context pairs, 
(iii) training a regression model to predict mean disagreement values. 

\subsection{Semantic Representations}
We employ the SentenceTransformer \textit{paraphrase-xlm-r-multilingual-v1} model to generate semantic embeddings for sentence pairs. This model is based on XLM-RoBERTa \cite{conneau-etal-2020-unsupervised}, a transformer architecture fine-tuned for multilingual sentence representation tasks. Given a context sentence, $C$, the embedding function $E(C)$ produces a 768-dimensional vector: 
\[
E(C) \in \mathbb{R}^{768}
\]
For each data sample, two contexts $C_1$ and $C_2$ are processed, and their embeddings are concatenated:
\[
X = [E(C_1), E(C_2)] \in \mathbb{R}^{1536}
\]

\subsection{Deep Regression Model}
We propose a deep feedforward neural network to map concatenated embeddings to mean disagreement scores. The model architecture consists of: Input Layer: 1536-dimensional concatenated embeddings. Hidden Layers: Four fully connected layers with dimensions [512, 256, 128, 64], each followed by BatchNorm and dropout ($p=0.3$). Output Layer: A single neuron for regression output.
Each hidden layer uses ReLU activation, and the loss function is Mean Squared Error (MSE):
\[
\mathcal{L} = \frac{1}{N} \sum_{i=1}^{N} \left( y_i - \hat{y}_i \right)^2
\]
where $y_i$ and $\hat{y}_i$ are the ground truth and predicted scores.
\begin{figure*}
  \includegraphics[width=\textwidth]{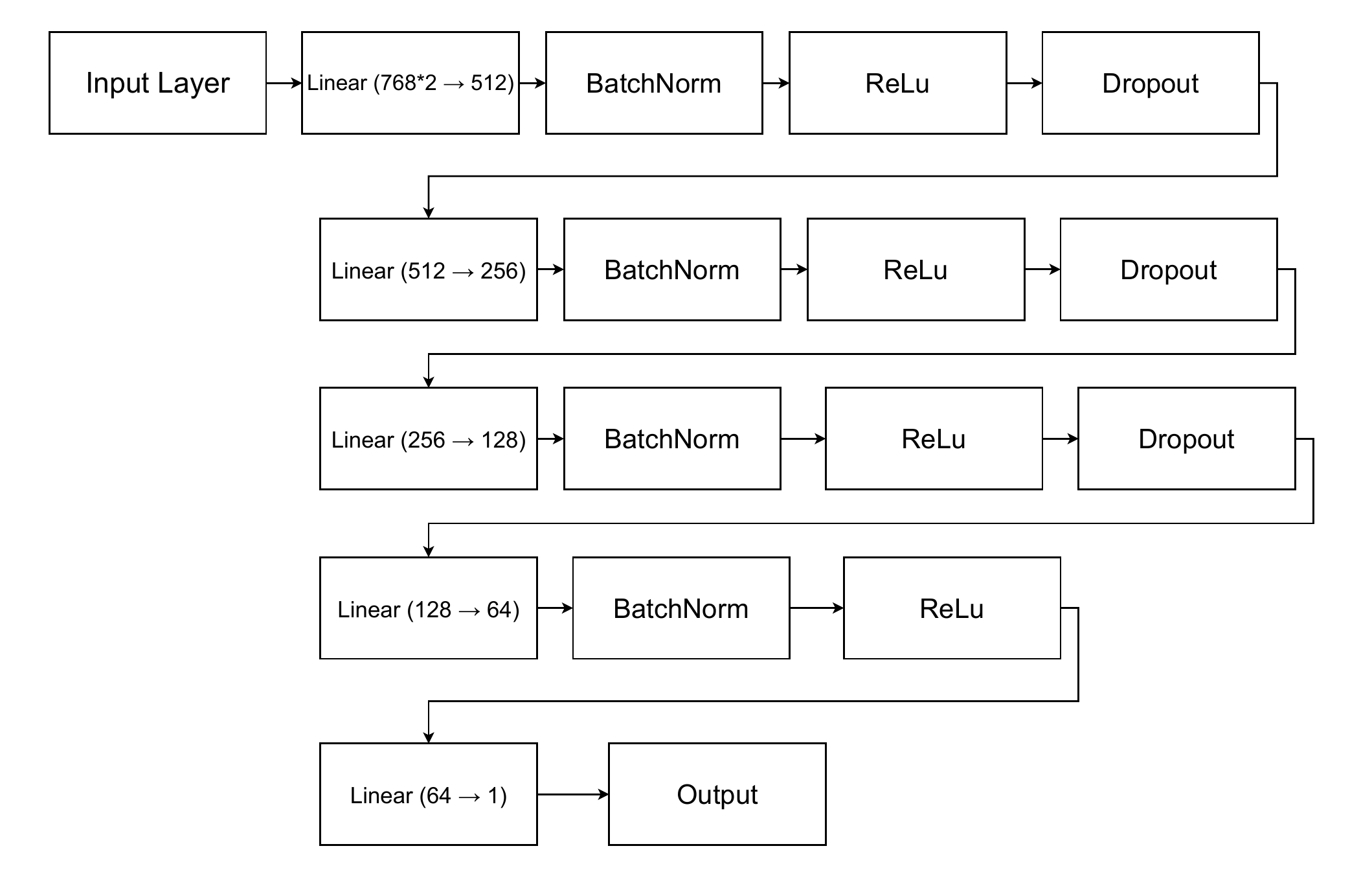}
  \caption{The structure of Deep Regression model.}
  \label{fig:deep}
\end{figure*}
\subsection{XLM-RoBERTa}
\begin{figure*}
  \centering
  \includegraphics[width=\textwidth]{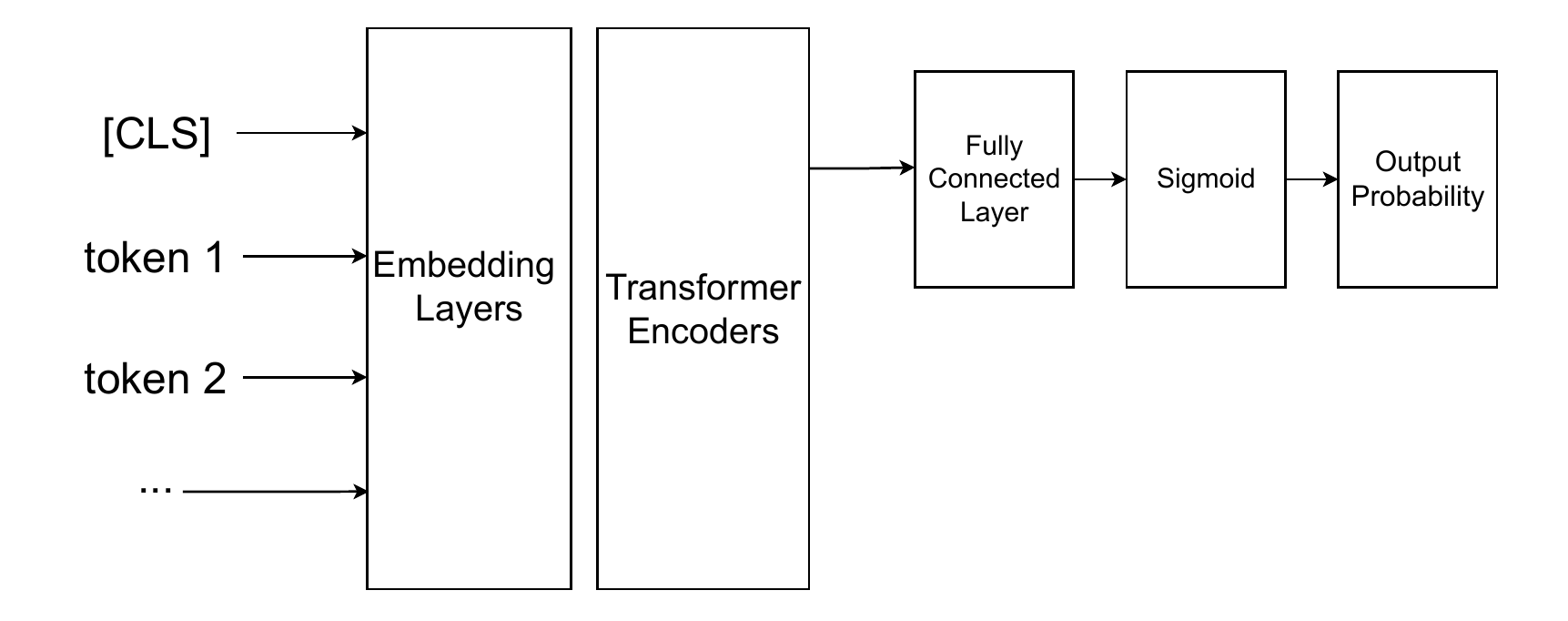}
  \caption{Structure of BERT and XLM-RoBERTa.}
  \label{fig:XLM}
\end{figure*}

As illustrated in Figure~\ref{fig:XLM}, the structure of XLM-RoBERTa \cite{conneau-etal-2020-unsupervised} consists of three main components: Embedding Layers, Transformer Encoders, and a final layer for handling specific tasks. During the model's training process, the input is a sequence of tokens, starting with the [CLS] character. The representation of the sequence is extracted from the vector C, corresponding to the [CLS] token. This vector is passed through a Fully Connected Layer and then processed using the sigmoid activation function to convert the output into a probability value. This value is optimized through the cross-entropy loss function.

\subsection{Training Strategy}
The model is trained using the AdamW optimizer with weight decay and an initial learning rate of $10^{-4}$. To prevent overfitting, we employ learning rate scheduling via \texttt{ReduceLROnPlateau}, reducing the learning rate by a factor of 0.5 if the validation loss does not improve for three consecutive epochs. Gradients are clipped \cite{NEURIPS2020_9ecff545} to a maximum norm for stability:
\[
\text{f}(g) = \min \left( 1, \frac{\text{max\_grad\_norm}}{\|g\|_2} \right) \cdot g
\]
\begin{figure}[H]
  \includegraphics[width=\columnwidth]{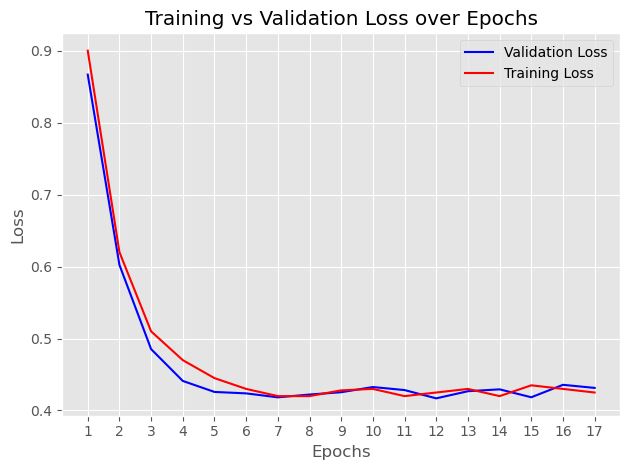}
  \caption{Training and validation loss while training.}
  \label{fig:experiments}
\end{figure}
\vspace{-35pt} 
\subsection{Evaluation Metrics}
The system's performance is evaluated using Spearman's Rank Correlation Coefficient ($\rho$) \cite{Zar2005SpearmanRC} between the predicted and true mean disagreement rankings. This metric is defined as:
\[
\rho = 1 - \frac{6 \sum_{i=1}^{N} d_i^2}{N(N^2 - 1)}
\]
where $d_i$ is the difference between the ranks of corresponding predicted and ground truth values, and $N$ is the total number of samples.
\begin{table*}[h]
    \centering
    \renewcommand{\arraystretch}{1.5} % Adjust row height
    \resizebox{\textwidth}{!}{ % Resize table to fit the page width
    \begin{tabular}{c|c|cccccccc}
        \hline
        \textbf{Phase} & \textbf{Team} & \multicolumn{8}{c}{\textbf{Subtask2 (spearman)}} \\ 
        \cline{3-10}
        & & \textbf{AVG} & \textbf{ZH} & \textbf{EN} & \textbf{DE} & \textbf{NO} & \textbf{RU} & \textbf{ES} & \textbf{SV} \\ 
        \hline
        \multirow{3}{*}{\textbf{Evaluation}} 
        & deep-change & \textbf{0.226 (1)} & 0.301 (7) & \textbf{0.078 (1)} & \textbf{0.204 (1)} & \textbf{0.286 (1) }&\textbf{ 0.175 (1)} & \textbf{0.187 (1)} &\textbf{ 0.350 (1)} \\ 
        \cline{2-10}
        & GRASP & 0.220 (2) &\textbf{ 0.539 (1)} & 0.042 (5) & 0.108 (2) & 0.272 (2) & 0.167 (2) & 0.115 (2) & 0.296 (2) \\ 
        \cline{2-10}
        &\textbf{ FuocChuVIP123 (ours)}& 0.124 (4) & 0.362 (4) & 0.018 (7) & 0.099 (3) & 0.156 (4) & 0.050 (6) & 0.012 (7) & 0.172 (3) \\ 
        \hdashline      
        \multirow{3}{*}{\textbf{Post evaluation}} 
        & deep-change & \textbf{0.281 (1)} & \textbf{0.574 (1)} & \textbf{0.143 (1)} &\textbf{ 0.241 (1)} & \textbf{0.294 (1) }&\textbf{ 0.194 (1)} & \textbf{0.161 (1)} & \textbf{0.360 (1)} \\ 
        \cline{2-10}
        & GRASP & 0.220 (2) & 0.539 (2) & 0.042 (3) & 0.108 (3) & 0.272 (2) & 0.167 (2) & 0.115 (2) & 0.296 (2) \\ 
        \cline{2-10}
        & funzac & 0.170 (3) & 0.433 (3) & 0.056 (2) & 0.167 (2) & 0.178 (3) & 0.076 (3) & 0.088 (3) & 0.194 (3) \\ 
        \hline
    \end{tabular}
    }
    \caption{Top 3 results of Subtask 2.}
    \label{tab:1}
\end{table*}
\section{Experimental setup}
For the shared task, we used a custom deep regression model built with a multi-layer perceptron (MLP) architecture, which was trained to predict mean disagreement scores from sentence embeddings. The embeddings were generated using the Sentence-Transformer model paraphrase-xlm-r-multilingual-v1, which was fine-tuned for multilingual text. We trained the model for 17 epochs with a batch size of 32 with PyTorch. The AdamW optimizer was used with an initial learning rate of 0.0001, and we applied a learning rate scheduler (ReduceLROnPlateau) with a patience of 3 epochs and a factor of 0.5 to reduce the learning rate when the validation loss plateaued. The model also utilized batch normalization and dropout layers to prevent overfitting.The training data was split into training and validation sets with an 80-20\% split. For evaluation, we used the mean squared error (MSE) loss for training and Spearman's rank correlation coefficient to assess the performance of the model.
Regarding data preprocessing, we used the raw contexts from the dataset without extensive cleaning. We merged the necessary information from training and development sets to construct the input for our model. No lemmatization or punctuation removal was applied as the dataset was in multilingual form, and we decided to focus on the context and target token indices for each pair of words.
Our model was evaluated on the development set, and we used Spearman’s rank correlation as the primary evaluation metric.

\section{Results}

Table 3 lists the evaluation phase scores of the top three contenders for subtask 2 as well as our
system. During this phase, submission scores and leaderboards were hidden. For Subtask 2, our team ranked 3rd out of 7 teams in the evaluation phase. We focused solely on Subtask 2 and did not participate in Subtask 1. The models of the top-performing teams utilized a variety of strategies. Our approach involved using embeddings generated from a pre-trained multilingual transformer model (XLM-R) to capture context information. These embeddings were then fed into a deep neural network model with batch normalization layers, which we trained to predict the "mean disagreement" score for each pair of contexts. We conducted a series of experiments with different hyperparameters and fine-tuned the model, which allowed us to achieve notable improvements in performance. In the evaluation phase, our team faced challenges, particularly with the Latin languagues, which proved to be more complex due to its size and variability. This likely contributed to our lower score of 0.124 on average  during the evaluation.
\section{Conclusion}
 In this paper, we presented our approach to Subtask 2 of the CoMeDi Shared Task, focusing on predicting disagreement rankings in multilingual word-in-context judgments. By leveraging sentence embeddings from the pre-trained paraphrase-xlm-r-multilingual-v1 model and a deep regression network with batch normalization, our method achieved competitive performance, ranking 3rd among 7 teams. Our results highlight the potential of multilingual embeddings and robust neural architectures for handling disagreement in semantic similarity tasks. Future work could explore further refinements to address language-specific complexities and improve overall model performance.
\section{Limitations}
Our system, while achieving competitive performance, has several limitations. First, it struggled with Latin-based languages like Spanish, highlighting challenges with XLM-RoBERTa embeddings for specific linguistic nuances. Second, the approach relied heavily on embedding quality, which may not fully capture fine-grained word-use differences. Additionally, the system focused solely on mean disagreement scores without modeling the underlying causes of annotator disagreement, such as cultural or subjective biases.

\bibliography{custom}
\end{document}